\documentclass{article}

\PassOptionsToPackage{numbers, compress}{natbib}

\usepackage[preprint]{neurips_2023}

\usepackage{comment}
\usepackage{amsthm}
\newtheorem{theorem}{Theorem}[section]

\newtheorem{lemma}[theorem]{Proposition}
\usepackage{titletoc}

\newcommand{\thistheoremname}{}
\newtheorem*{genericthm*}{\thistheoremname}
\newenvironment{namedthm*}[1]
  {\renewcommand{\thistheoremname}{#1}%
   \begin{genericthm*}}
  {\end{genericthm*}}

\usepackage[utf8]{inputenc} 
\usepackage[T1]{fontenc}    
\usepackage{hyperref}       
\usepackage{url}            
\usepackage{booktabs}       
\usepackage{amsfonts}       
\usepackage{nicefrac}       
\usepackage{microtype}      
\usepackage{xcolor}         

\usepackage{amsmath}               
  {
      
  }
\usepackage{multirow}
\usepackage{makecell}
\usepackage{graphicx}
\usepackage{subcaption}
\usepackage{algorithm}
\usepackage{algpseudocode}
\usepackage{wrapfig}
\usepackage{listings}
\DeclareMathOperator*{\argmin}{arg\,min}

\newtheorem{definition}{Definition}[section]

\newcommand{\E}{\mathbb{E}}
\newcommand{\V}{\mathbb{V}ar}
\renewcommand{\P}{\mathbb{P}}

\title{DBsurf: A Discrepancy Based Method for Discrete Stochastic Gradient Estimation}

\author{%
  Pau Mulet Arabi\thanks{equal contribution - work completed while interns at Sony AI} \\
  EPFL, Switzerland\\
  \texttt{pau.muletarabi@epfl.ch} \\
  \And
  Alec Flowers\footnotemark[1]\\
  EPFL, Switzerland\\
  \texttt{alec.flowers@epfl.ch}\\
    \And
  Lukas Mauch\\
  Sony Europe B.V., Germany\\
  \texttt{lukas.mauch@sony.com}\\
  \And
  Fabien Cardinaux\\
  Sony Europe B.V., Germany\\
  \texttt{fabien.cardinaux@sony.com}\\
}

\begin{document}

\maketitle

\begin{abstract}
Computing gradients of an expectation with respect to the distributional parameters of a discrete distribution is a problem arising in many fields of science and engineering. Typically, this problem is tackled using Reinforce, which frames the problem of gradient estimation as a Monte Carlo simulation. Unfortunately, the Reinforce estimator is especially sensitive to discrepancies between the true probability distribution and the drawn samples, a common issue in low sampling regimes that results in inaccurate gradient estimates. In this paper, we introduce DBsurf, a reinforce-based estimator for discrete distributions that uses a novel sampling procedure to reduce the discrepancy between the samples and the actual distribution. To assess the performance of our estimator, we subject it to a diverse set of tasks. Among existing estimators, DBsurf attains the lowest variance in a least squares problem commonly used in the literature for benchmarking. Furthermore, DBsurf achieves the best results for training variational auto-encoders (VAE) across different datasets and sampling setups. Finally, we apply DBsurf to build a simple and efficient Neural Architecture Search (NAS) algorithm with state-of-the-art performance.
\end{abstract}

\section{Introduction}
The optimization over a parametric family of probability distributions where the objective function takes the form of an expectation with respect to a given probability distribution is an important component to solving problems in statistical inference, machine learning, sensitivity analysis, queuing theory, and experimental design \cite{JMLR:v21:19-346}. In general, the problem can be formulated as the minimization of an objective function $\mathcal{L}: \Theta \mapsto \mathbb{R}$ of the form 
\begin{equation}
    \mathcal{L}(\theta) = \mathbb{E}_{x\sim p_\theta}[f(x)] \label{eq:1}
\end{equation} 
where \(\Theta\) is a connected subset of \(\mathbb{R}^d\) and \(\{p_\theta: \theta \in \Theta \}\) is a parametric family of probability distributions.

\begin{figure}[h]
\vspace{-20pt}
    \centering
    \includegraphics[width=1\columnwidth]{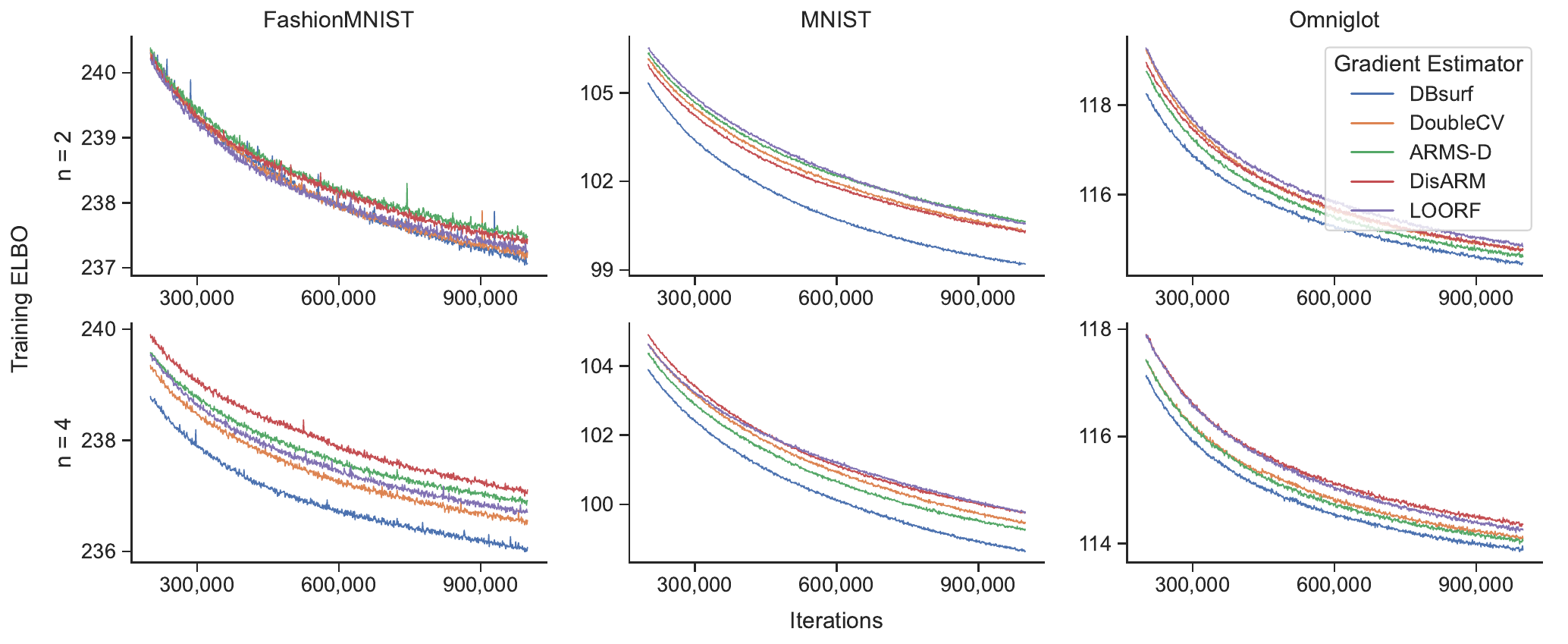}
    \caption{Training a non-linear binary VAE on FashionMNIST, MNIST and Omniglot. The average training ELBO over 5 seeds is plotted by iterations for each dataset. DBsurf outperforms the other discrete gradient estimators. \textit{Top:} $n=2$ samples. \textit{Bottom: } $n=4$ samples.}
    \label{fig:training_vae}
\end{figure}

In the field of machine learning this problem appears in model-free reinforcement learning \cite{Sutton1998}, optimizing variational auto-encoders, \cite{Kingma2013AutoEncodingVB} and in neural architecture search \cite{cai2018proxylessnas}. The standard way to tackle \eqref{eq:1} is to use some sort of gradient based optimization method. In the discrete case, we have an analytic expression for the gradient of the objective 
\begin{equation}\label{eq:2}
    \nabla \mathcal{L}(\theta) = \hspace{-8pt}\sum_{x \in supp(p_\theta)}\hspace{-8pt} p_\theta(x)f(x) \nabla_{\theta}\log p_\theta(x)
\end{equation}
which is in most cases computationally intractable because of the complexity of evaluating \(f\) at many points. In practice, we need to resort to numerical estimates.

When the random variables with respect to which the expectation is taken are continuous, the gradients can often be successfully estimated using the Reparametrization Trick \cite{Kingma2013AutoEncodingVB} or one of its extensions \cite{Ruiz2016TheGR}, \cite{figurnov2018implicit}. However, in the discrete setting, it is less clear how to conduct the estimation.  An initial successful approach that worked for discrete distributions was Reinforce \cite{Williams1992SimpleSG}. It reduced the problem of estimating the gradient of an expectation to a classical Monte-Carlo estimation of \eqref{eq:2}. 

The Reinforce estimator replaces the true probabilities $p_\theta(x)$ in \eqref{eq:2} by the empirical ones $\hat{p}(x)$. Consequently, if we use i.i.d samples we get a consistent estimator since $\hat{p}$ is guaranteed to converge to $p$ as the $n$, the number of samples, goes to infinity. Unfortunately, that convergence is slow, $\mathcal{O}(\frac{1}{\sqrt{n}})$, so in low sampling regimes the Reinforce estimator tends to give inaccurate gradients.
In this paper, we address that problem by introducing a novel sampling strategy for stochastic gradient estimation.

Our contributions are:
\begin{itemize}
    \item We introduce a new low-discrepancy sampling technique DBsample for discrete distributions in Sec. \ref{section:method} and use it to build a Reinforce-based gradient estimator DBsurf. By modifying the distribution at each step, we get more representative samples of the actual distribution (lower discrepancy), leading to more accurate gradient estimates. Indeed, we show that our estimator outperforms all the existing estimators for training discrete variational autoencoders across multiple datasets and training setups (Fig. \ref{fig:training_vae}).
    \item In Sec. \ref{section:nas} we identify the important problem of Neural Architecture Search (NAS) where discrete sampling techniques can be used. We explain why optimizing with a discrete distribution is preferable in the NAS context, giving further justification for the importance of creating better gradient estimators for discrete distributions. Motivated by this, we leverage our estimator to build an efficient method for Neural Architecture Search and show that it achieves state of the art results.
\end{itemize}
\vspace{-10pt}
\section{Method}\vspace{5pt}
\label{section:method}
\subsection{Background}
Let $\textbf{x} = \{x_1, x_2, \dots, x_n\}$ be a set of $n$ samples with $x_i\in \mathbb{R}^d$ such that $x_i \sim p_\theta(x_i) = \text{Categorical}(\sigma(\theta_i))$, where $\sigma(\theta_j) = \frac{\exp{\theta_j}}{\sum_{k=1}^d \exp{\theta_k}}$. For the random variables, we use subscripts for sample indices and superscripts for the dimension over the categorical variable while for parameters we only use subindices (to refer to the dimension).

The most direct way to estimate \eqref{eq:2} is to perform a Monte-Carlo simulation
\begin{equation}
    \label{eq:3}
    g_{\scalebox{0.5}{RF}}(\textbf{x}) = \frac{1}{n}\sum_{i=1}^n f(x_i) \nabla_\theta \log p_\theta(x_i)
\end{equation}
which is the Reinforce estimator proposed in \cite{Williams1992SimpleSG}. This estimator is unbiased, but it suffers from having high variance. A popular approach to reduce the variance are control variates \cite{Robert2005MonteCS}. In the particular case of \eqref{eq:3}, we can take advantage that the score has zero mean \(\mathbb{E}[\nabla_\theta \log p_\theta(x)] = 0\) to build control variates with it. This leads to the general class of estimators
\begin{equation}
g(\textbf{x}) = \frac{1}{n}\sum_{i=1}^n (f(x_i)-z_i) \nabla_\theta \log p_\theta(x_i)
\end{equation}
where \(z_i\) are random variables independent of \(x_i\) (for the same index) so that the estimator remains unbiased. In principle, one could take \(z\) from an arbitrary distribution, but to avoid any computational overhead, it is common to generate \(\mathbf{z}\) from the already drawn sample set \(\mathbf{x}\). A successful example of this strategy, proposed in \cite{Kool2019Buy4R}, is Leave-one-out Reinforce (LOORF)
\begin{equation}
\label{eq:5}
g_{\scalebox{0.5}{LOORF}}(\textbf{x}) = \frac{1}{n}\sum_{i=1}^n \left(f(x_i)-\frac{1}{n-1}\sum_{j\neq i}f(x_j)\right) \nabla_\theta \log p_\theta(x_i)
\end{equation}
that corresponds to \(z_i = \frac{1}{n-1}\sum_{j\neq i}f(x_j)\) which is independent of \(\nabla_\theta \log p_\theta(x_i)\) when the different samples \(x_1,...,x_n\) are independent. 

\subsection{The Importance of Discrepancy}
\begin{definition}(Discrepancy)
    Let $\textbf{x}=\{x_1,...,x_n\}$ be a set of samples and let $p$ be a categorical distribution. We define the discrepancy of the set $\textbf{x}$ with respect to the probability distribution $p$ as:
    \begin{equation}
    D_p(\textbf{x})=\hat{p}-p=\frac{1}{n}\sum_{i=1}^{n}x_i-p
\end{equation}
\end{definition}

where \(\hat{p}\) denotes the empirical probability \(\hat{p}=\frac{1}{n}\sum_{i=1}^{n}x_i\). This definition of discrepancy is a natural extension for categorical distributions which differs slightly from the classical definition for uniform distributions in the context of Quasi Monte Carlo Methods \cite{niederreiter1992random}.

We take LOORF \eqref{eq:5} as the baseline for our estimator and focus on the sampling procedure to further improve its performance. In the particular case of a categorical distribution with a softmax parametrization, the estimator can be written as
\begin{equation}
\label{eq:6}
g_{\scalebox{0.5}{LOORF}}(\textbf{x}) = \underbrace{\frac{1}{n-1}\sum_{i=1}^{n}f(x_i)(x_i-p)}_{\text{Reinforce \footnotemark }} - \underbrace{(\hat{p}-p)\frac{1}{n-1}\sum_{i=1}^{n}f(x_i)}_{\text{Discrepancy correction}}
\end{equation}\footnotetext{\vspace{-22pt}Slight abuse of terminology since the unbiased version of reinforce should be $1/n$}

The difference between LOORF and the classical Reinforce can be seen just as a correcting term to compensate for the inconsistencies between the samples and the true distribution \eqref{eq:6} (Appendix \ref{appendix:theory}). Given the performance gap \cite{Kool2019Buy4R} between Reinforce and LOORF we can expect the discrepancy between the samples and the true distribution to play a crucial role in the success of an estimator.

We motivate our method by demonstrating a source of inaccurate gradients when using Monte-Carlo estimation. 
In the following example, we show that the discrepancy between the samples and the underlying true probability can lead to bad gradient updates. 

Consider the following least-squares problem in two dimensions where $\phi(\theta) = \frac{1}{1 + e^{-\theta}}$ and $\theta$ is initialized such that $p_\theta$ is $0.5$. Two samples $x_1, x_2$ are drawn from $p_\theta$. Note the optimum is $p_\theta$ equals $1$ as this maximizes the least squared expression.
\begin{gather*}
    \max_{\theta \in \Theta}\mathbb{E}_{x \sim p_\theta(x)}[\Vert x - 0.49 \Vert^2] \quad \text{where } p_\theta \sim \text{Bernoulli}(\phi(\theta))\\ \vspace{15pt}
    p_\theta = \Bigl(\substack{$0.5$\\$0.5$}\Bigl) \quad
    x_1 = \Bigl(\substack{$0$\\$1$}\Bigl) \quad x_2 = \Bigl(\substack{$1$\\$1$}\Bigl) \quad \hat{p} = \Bigl(\substack{$0.5$\\$1.0$}\Bigl)\vspace{15pt}
\end{gather*}
In the first dimension, there is no discrepancy, as $\hat{p}_1$ equals $(p_\theta)_1$. But in the second dimension, there is high discrepancy.
   $$D_{p_\theta}(x_1, x_2) = \hat{p}-p_\theta = \Bigl(\substack{$0$\\$0.5$}\Bigl)$$

\begin{minipage}{0.48\textwidth}
   
In Fig. \ref{fig:toy_gradients} the true gradient, the Reinforce estimate and the LOORF estimate are plotted. In the axis of low discrepancy, reinforce has the proper gradient magnitude and direction. However, in the axis of high discrepancy, the Reinforce gradient is pointing in the wrong direction. LOORF corrects this discrepancy by shrinking in the direction of high discrepancy and pushing in the direction of low discrepancy.  \\

We have seen that if the samples drawn from the distribution do not represent the underlying parameter, the gradient estimates are significantly deteriorated. 
In order to circumvent this problem we design a novel sampling procedure that helps us avoid drawing samples that lead to high discrepancy. 

\end{minipage}\hfill 
\begin{minipage}{0.50\textwidth}
\vspace{-20pt}
    \includegraphics[width=\textwidth]{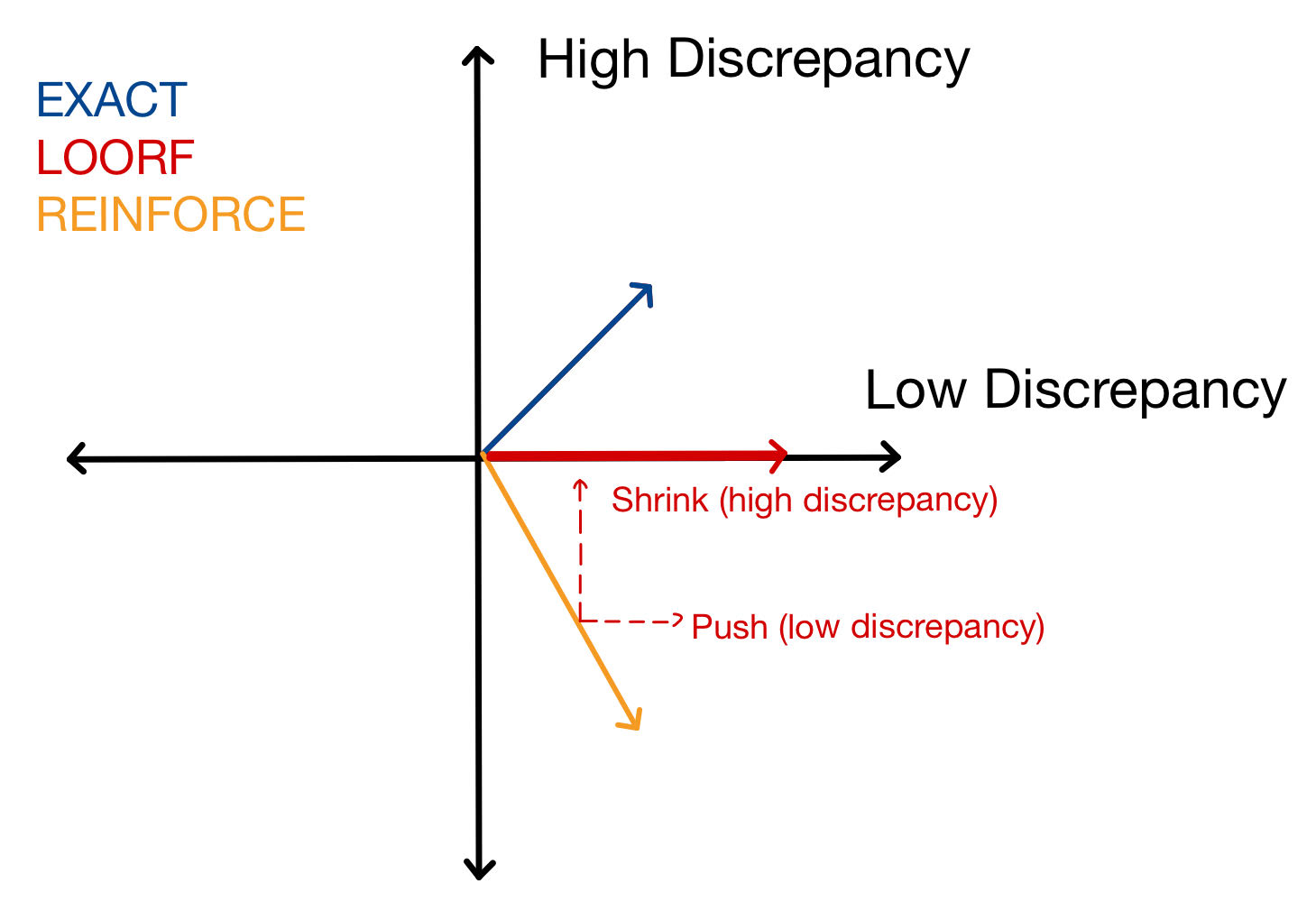}\\
    \hspace{10pt }Figure 2: Toy example gradients
    \label{fig:toy_gradients}
\end{minipage}

\subsection{Discrepancy Based Sampling}
We introduce our algorithm and theoretical results for Bernoulli random variables as this is the main setting where we compare against other gradient estimators. In \ref{section:general_discrete}, we discuss how sampling from a general discrete distribution can be reduced to this setting, and in \ref{section:nas} apply our sampling to a categorical distribution.

\begin{figure}[h]
\vspace{-10pt}
    \centering
    \includegraphics[width=1.0\columnwidth]{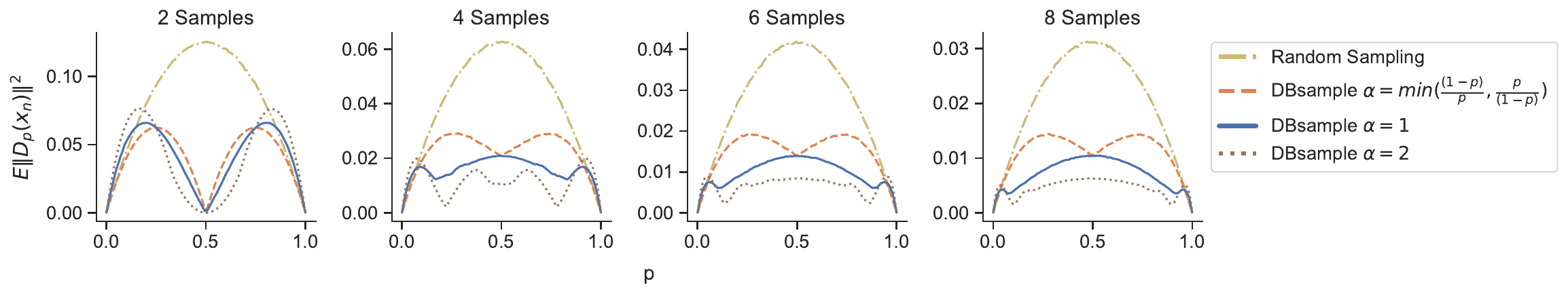}
    \caption{The mean squared error for discrepancy $\mathbb{E}[\Vert D_p(\textbf{x})\Vert^2]$ plotted over $p$ for $n = \{2, 4, 6, 8\}$.}
    \label{fig:less_discrep}
\end{figure}

\label{alg:algorithm}
        \begin{algorithm}[H]
            \caption{Discrepancy-Based Sampling (DBsample)}
            \begin{algorithmic}
             
                \State \textbf{Input: } \(p\) (true probabilities), \(n\) (number of samples), \(\alpha\) (correction factor) 
                \State \vspace{-5pt}
                \State $\hat{p}$ $\gets$ $p$
                \State $q$ $\gets$ $p$
                
                \For{\(i=1,...,n\)}
                    \State \(x_i \sim Bernoulli(q)\)
                    \State $\hat{p} = ((i-1) \cdot \hat{p} + x_i)/i$
                    \State \(q = \min(1, (p (1+\alpha) - \alpha \hat{p})_{+})\)
                \EndFor
                \State \vspace{-5pt}
                \State \textbf{Output: }\(\{x_1,...,x_n\}\)
            \end{algorithmic}
        \end{algorithm}
The main objective of our strategy is to correct the discrepancy between the samples and the actual distribution. We achieve this by modifying the distribution at each step based on the previously drawn samples. This approach generates samples that are more consistent  with the true distribution (lower discrepancy) than the commonly used i.i.d sampling.
\begin{lemma} \label{lemma:1} Let $\textbf{x}=\{x_1,...,x_n\}$ be a set of i.i.d random variables with distribution $Bernoulli(p)$ for arbitrary $p \in (0, 1)^d$ and $n \geq 2$. Consider the set  $\textbf{z}=\{z_1,...,z_n\}$ of random variables associated to Algorithm 1 with initial parameters $p, n$ and $\alpha \leq \min \left( \frac{1-p_k}{p_k}, \frac{p_k}{1-p_k}\right) \;$ for any k. We have:
$$
\mathbb{E}[\|D_p(\textbf{z})\|^2] < \mathbb{E}[\|D_p(\textbf{x})\|^2]
$$
\end{lemma}
The previous proposition provides the guarantee that samples generated by Algorithm 1 will have (in expectation) lower discrepancy than i.i.d random samples. It is a qualitative result covering many ranges of the parameters. In Fig. \ref{fig:less_discrep} we can see how the discrepancy of DBsample varies for different choices of the parameters and how it compares to random i.i.d sampling.

\begin{figure}[h]
    \centering
    \includegraphics[width=.9\columnwidth]{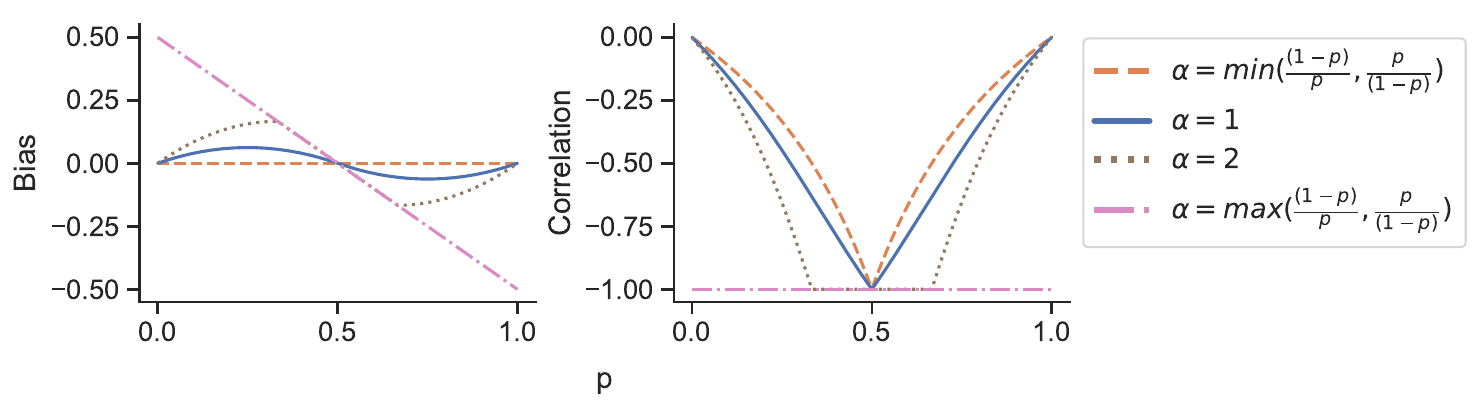}
    \caption{Adjusting the $\alpha$ parameter in DBsample allows one to trade off bias for negative correlation. The bias and correlation are shown for different $\alpha$ where $d=1$ and $n=2$ of our estimation problem (\(\nabla \mathbb{E}_p[f]\)). \textit{Left:} Bias is measured as \(\frac{\mathbb{E}[X_1 + X_2]}{2} - p\). When \(\alpha = \min (\frac{1-p}{p}, \frac{p}{1-p})\) the probability distribution is preserved and as we increase $\alpha$ we slowly add bias that pushes the distribution towards the center. \textit{Right:} Correlation between the two samples $(x_1, x_2)$. When $\alpha = \min (\frac{1-p}{p}, \frac{p}{1-p})$ the samples are negatively correlated, but by breaking the identically distributed assumption we can get more negatively correlated samples. 
    }
    \label{fig:bias_var_dbsample}
\end{figure}

Intuitively, DBsample reduces the discrepancy by modifying the probability of a given category based on whether it is over-represented or under-represented in the current population (according to the true distribution). The parameter \(\alpha\) determines the magnitude at which this happens. We can even change the marginal distribution if we strongly favor the discrepancy correction. However, if one does not want this behavior, it suffices to take the parameter within certain bounds.
\begin{lemma} \label{lemma:2} For any \(p \in [0, 1]^d\) and any \(n\in \mathbb{N}\) if we take $
\alpha \leq \min \left( \frac{1-p_k}{p_k}, \frac{p_k}{1-p_k}\right)
$ for any k, the sampling procedure in Algorithm 1 preserves the original distribution.  Equivalently $$\E[x_k] = p \quad  k=1,...,n$$
\end{lemma}
The previous result provides guarantees to preserve the original distribution when sampling with our algorithm, but we don't need to enforce this. In most of our experiments, we observed that values of \(\alpha\) above the threshold led to the best performance. The intuition for this is that preserving the true distribution is very beneficial to get consistent estimators when approaching an asymptotic regime. However, in low sampling experiments, it is better to seek that "consistency" directly. In Fig. \ref{fig:bias_var_dbsample} we can see the trade-off between modifying the probability distribution by introducing bias and generating more negatively correlated samples which as shown in Fig. \ref{fig:less_discrep} helps with discrepancy. 
Proofs of Prop. \ref{lemma:1} and Prop. \ref{lemma:2} are in Appendix \ref{appendix:theory}

 In order to better understand the role of the marginal distributions and the correlation structure in the performance of the estimator, let's consider the simple case of a Bernoulli distribution
with dimension \(d=1\) and \(n=2\) samples. We seek to estimate (\(\nabla \mathbb{E}_p[f]\)) for an arbitrary \(p \in (0, 1)\) and an arbitrary function \(f: \{0, 1\} \mapsto \mathbb{R}\).  The true gradient of \(\mathbb{E}_p[f]\) is given by \(\nabla \mathbb{E}_p[f] = (f(1)-f(0))p(1-p)\) while we consider the general class of estimators
\begin{equation}
    g_q(x_1, x_2) = \frac{p(1-p)}{\kappa_q}(f(x_1)-f(x_2))(x_1-x_2)  \qquad (x_1, x_2) \sim q
\end{equation}
and want to see what is the role of the distribution \(q\) in \(\{0, 1\}^2\), from which we generate the samples in the variance of the estimator \(g\). To make the variance comparison fair we consider unbiased versions by taking \(\kappa_q = q(x_1=1, x_2=0)+q(x_1=0, x_2=1)\). A quick computation (Appendix \ref{appendix:theory}) yields the dependence of the variance on the correction parameter:
\[
\mathbb{V}ar[g_q(x_1, x_2)] =
\begin{cases}
\label{eq:var_g}
C_f \:p(1-p)[\frac{1}{2}-p(1-p)] & \alpha = 0 \quad \text{i.i.d}\\
C_f \: p(1-p)[\frac{1}{2(1+\alpha)}-p(1-p)] & \alpha \leq \min \left( \frac{1-p}{p}, \frac{p}{1-p}\right) \\ 
C_f \: p(1-p)[\frac{\max(p, 1-p)}{\max(p, 1-p)(1+\alpha)+1}-p(1-p)] & \alpha \leq \max \left( \frac{1-p}{p}, \frac{p}{1-p}\right)\\
0 & \alpha \geq \max \left( \frac{1-p}{p}, \frac{p}{1-p}\right)
\end{cases}
\]
where $C_f$ is a constant depending on the function. This shows the role of the correction parameter in the estimator's performance and why it is not the best option to set it within the boundaries that make the samples preserve the distribution. The sampling procedure  generally has three possible regimes we can distinguish: start with independent and identically distributed samples \(\alpha=0\), which leads to the highest variance; increase \(\alpha\) while preserving the distribution, resulting in higher dependence and consequently variance reduction; reach a point where \(\alpha\) cannot be further increased without modifying the distribution (neither can the dependence), but which further reduces the variance. In this particular example, by increasing \(\alpha\) enough we recover the true gradient, but for higher dimensions the evaluations required to compute the true gradient are usually much larger than the number of samples.

\begin{figure}[h]
    \centering
    \includegraphics[width=1\columnwidth]{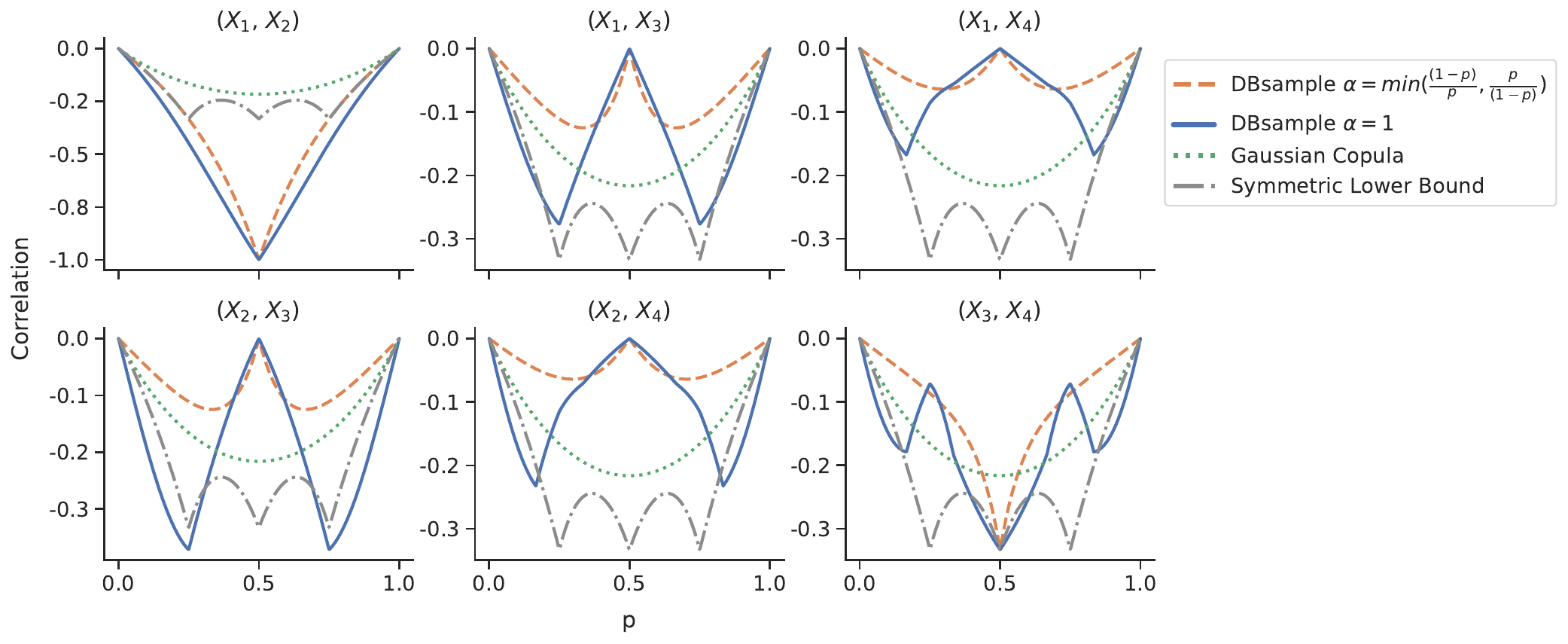}
    \caption{Correlation between 4 samples $\{x_1, x_2, x_3, x_4\}$ for all probabilities $p$ for DBsample with two different $\alpha$ parameters, a gaussian copula \cite{Dimitriev2021ARMSAG}, and a lower bound on the minimum correlation between symmetrically distributed bernoulli random variables \cite{HOERNIG2018469}.}
    \label{fig:corr}
\end{figure}

While the motivation for our sampling procedure was to reduce discrepancy, as a side effect, DBsample induces negative correlation between samples similar to that of antithetic sampling.
Antithetic samples can reduce variance when the correlation between samples is negative such that $\text{Corr}(f(\textbf{x}) \nabla_\theta p_\theta(\textbf{x}), f(\textbf{x}_{\text{anti}}) \nabla_\theta p_\theta(\textbf{x}_{\text{anti}})) < 0$ where $\textbf{x}_{\text{anti}}$ are antithetic samples built from $\textbf{x}$ normal samples \cite{owenbook}.
A recent approach to improving discrete gradient estimators uses copulas to negatively correlate multiple samples from a Bernoulli distribution \cite{Dimitriev2021ARMSAG}.
We calculate the correlation structure (Appendix \ref{appendix:theory}) for DBsample and can directly compare.
As seen in Fig. \ref{fig:corr} DBsample $\alpha = 1$ produces more negatively correlated samples and this translates to better variance reduction over the copula approach on a toy problem Fig. \ref{fig:toy_variance}.
\vspace{-15pt}
\subsection{General Discrete Distributions}
\label{section:general_discrete}
In the previous section we introduced our sampling procedure in the context of Bernoulli random variables but, although our target applications use Bernoulli and categorical random variables, the algorithm can be used for arbitrary discrete distributions as far as we have independence across the dimensions. In the case of categorical distributions \hyperref[alg:algorithm]{Algorithm 1} works exactly in the same way, just noting that $\hat{p}$ and $x_i$ will be matrices instead of vectors. The clipping \(q = \min(1, (p (1+\alpha) - \alpha \hat{p})_{+})\) is applied to each category and then $q$ is normalized \(q \leftarrow q/sum(q)\). If we have a general discrete distribution the idea is to one-hot encode the support defining a categorical distribution, sample using the algorithm, and map back to the original support. More details can be found in Appendix \ref{appendix:general_discrete}.

\subsection{Related Work}
Even though the original Reinforce method has too much variance to make it useful in practice, the reformulation as a Monte Carlo estimation opened a path for developing many successful techniques.
Most importantly, it allowed the migration of variance reduction techniques from the field of Monte Carlo \cite{Asmussen2007StochasticS} to gradient estimation.
Applying these variance reduction techniques on top of the reinforce estimator has led to many successful gradient estimation methods.
For instance, antithetic variables arise in \cite{Yin2018ARMAG}, \cite{disarm} which use negatively correlated pairs of samples to reduce the variance of the gradient estimation. The method is further extended to an arbitrary number of correlated samples in \cite{Dimitriev2021ARMSAG} by using copulas to generate groups of mutually correlated uniform random variables that are then transformed into Bernoullis inheriting the correlation structure. Additionally, control variates have also been a crucial resource for the development of several good gradient estimators such as \cite{Tucker2017REBARLU} which uses a control variate on top of a continuous relaxation using the Gumbel-Softmax distribution \cite{jang2017categorical} \cite{maddison2017the} or \cite{Kool2019Buy4R} which makes a leave-one-out transformation to Reinforce, leading to considerable variance reduction. The latter has been further extended to an extra control variate using either continuous extensions of the estimated gradient \cite{Titsias2021DoubleCV} or Stein operators \cite{Shi2022GradientEW}.

Another line of work has leveraged the seminal Rao-Blackwell Theory \cite{rao1945information}, \cite{Blackwell1947ConditionalEA} to construct low variance gradient estimators \cite{liu2018darts}, \cite{Kool2020EstimatingGF}, \cite{NIPS2015_1373b284} as  the theory provides guarantees to reduce the variance of a general estimator by conditioning on sufficient statistics.

\section{Experiments}
We call our new estimator DBsurf, which is the combination of our sampling technique DBsample and LOORF. For all the following experiments we set $\alpha = 1$ as this breaks the identically distributed paradigm and shows a good trade-off between bias, negative correlation, and discrepancy correction.

\subsection{Toy Experiment}
We use the toy experiment introduced in \cite{tucker2017rebar} to compare the variance of the various gradient estimation techniques. This is also the same setup as our motivating example in the introduction
\begin{gather}
    \max_{\theta \in \Theta}\mathbb{E}_{x \sim p_\theta(x)}[\Vert x - 0.49 \Vert^2] \quad \text{where } p_\theta \sim \text{Bernoulli}(\phi(\theta))
\end{gather}
For the simplest case of a 1D problem, we can compute the debias term in closed form for our sampling technique (Appendix \ref{appendix:theory}). In this case we are interested in comparing the variance of our estimator against the base estimator LOORF, ARMS-D (which uses a Dirichlet copula to generate symmetrically correlated samples) and DoubleCV (an estimator that uses two control variates and optimizes a regression coefficient to reduce variance). 

We run the experiment for $n = {2, 4, 6, 8}$ samples, and for every value of $p_\theta$ we generate 1000 gradient estimates and compute the variance. For DoubleCV we optimize the regression coefficient $\beta$ on the previous value of $p_\theta$.

\begin{figure}[h]
    \centering    \includegraphics[width=1\columnwidth]{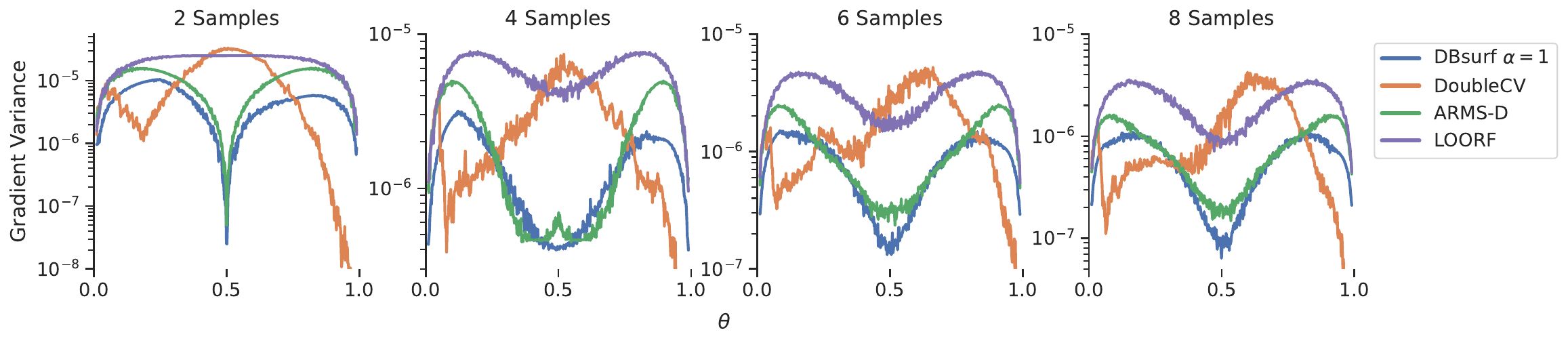}
    \caption{Gradient variance plotted on a log scale as $p_\theta$ ranges from $0$ to $1$.}
    \label{fig:toy_variance}
\end{figure}

For all samples DBsurf significantly reduces variance over the baseline of LOORF as shown in Fig. \ref{fig:toy_variance}. 
For two samples breaking the IID assumption indeed pays off and DBsurf $\alpha = 1$ reduces variance over ARMS-D.
As the number of samples gets larger, DBsurf $\alpha = 1$ has a lower variance in the tails and in the middle of the probability distribution compared to ARMS-D. 
In the middle of the distribution, DBsurf has lower variance than DoubleCV, but in the tails DoubleCV has better variance reduction due to the help of the regression coefficient $\beta$ which updates through training.

\subsection{Training Variational Auto-Encoders}
We consider the problem of training a variational auto-encoder \cite{Kingma2013AutoEncodingVB} where the variational distribution $q_{\phi}(z|x)$ is restricted to be a factorized Bernoulli distribution. We simultaneously learn an inference network $q_{\phi}(z|x)$ which approximates the true, but intractable posterior $p(z|x)$ and a generative network $p_{\theta}(x|z)$ by maximizing the variational lower bound (ELBO). 
This task is used as a benchmark for discrete gradient estimation techniques.

We use FashionMNIST \cite{fashionmnist}, MNIST \cite{mnist}, and Omniglot \cite{omniglot} datasets and compare against 4 other gradient estimators (DoubleCV, ARMS-D, DisARM and LOORF) which are designed to reduce variance when estimating gradients of expectations over discrete distributions. Our training setup is described in Appendix \ref{appendix:vae}
As shown in Table \ref{table:1}, DBsurf led to the lowest average training ELBO across all datasets and number of samples. By breaking the identically distributed assumption we can improve even in the 2 sample case. This result also highlights the benefits of focusing on discrepancy. All the estimators here are LOORF-based estimators which can be seen as Reinforce plus a term that corrects for the discrepancy, as shown in Equation \ref{eq:6}. By putting the most emphasis on discrepancy DBsurf is able to outperform the other estimators. 

\begin{table}[h]
\small
\centering
\caption{The mean $\pm$ standard error over 5 seeds of the final training ELBO for various gradient estimators. The method achieving the lowest average training ELBO is \textbf{bolded}.}
\label{table:1}
\begin{tabular}{@{}lllllll@{}}
\toprule
\multicolumn{1}{c}{\textbf{Dataset}} & \multicolumn{1}{c}{\textbf{n}} & \multicolumn{1}{c}{\textbf{DBsurf}} & \multicolumn{1}{c}{\textbf{DoubleCV}} & \multicolumn{1}{c}{\textbf{ARMS-D}} & \multicolumn{1}{c}{\textbf{DisARM}} & \multicolumn{1}{c}{\textbf{LOORF}} \\ \midrule
\multirow{4}{*}{FMNIST} & 2 & \boldmath{$237.07\pm0.28$} & $237.24\pm0.04$ & $237.49\pm0.27$ & $237.44\pm0.13$ & $237.26\pm0.16$ \\
 & 4 & \boldmath{$236.08\pm0.14$} & $236.56\pm0.26$ & $236.91\pm0.09$ & $237.11\pm0.17$ & $236.71\pm0.21$ \\
 & 6 & \boldmath{$235.32\pm0.24$} & $236.66\pm0.09$ & $237.02\pm0.11$ & $236.75\pm0.18$ & $236.94\pm0.17$ \\
 & 8 & \boldmath{$235.66\pm0.26$} & $236.63\pm0.22$ & $236.36\pm0.17$ & $236.55\pm0.29$ & $236.66\pm0.24$ \\
 &  &  &  &  &  &  \\
\multirow{4}{*}{MNIST} & 2 & \boldmath{$99.20\pm0.16$} & $100.31\pm0.19$ & $100.64\pm0.40$ & $100.33\pm0.42$ & $100.56\pm0.22$ \\
 & 4 & \boldmath{$98.66\pm0.25$} & $99.49\pm0.21$ & $99.29\pm0.15$ & $99.76\pm0.30$ & $99.79\pm0.08$ \\
 & 6 & \boldmath{$98.29\pm0.17$} & $99.53\pm0.33$ & $99.15\pm0.46$ & $99.58\pm0.25$ & $99.10\pm0.15$ \\
 & 8 & \boldmath{$98.56\pm0.29$} & $99.20\pm0.49$ & $98.79\pm0.35$
 & $99.41\pm0.20$ & $99.69\pm0.47$ \\
 &  &  &  &  &  &  \\
\multirow{4}{*}{Omniglot} & 2 & \boldmath{$114.47\pm0.08$} & $114.76\pm0.14$ & $114.61\pm0.08$ & $114.77\pm0.17$ & $114.83\pm0.16$ \\
 & 4 & \boldmath{$113.88\pm0.12$} & $114.09\pm0.21$ & $114.05\pm0.13$ & $114.36\pm0.15$ & $114.25\pm0.20$ \\
 & 6 & \boldmath{$113.59\pm0.06$} & $113.60\pm0.09$ & $113.63\pm0.15$ & $113.93\pm0.16$ & $114.01\pm0.17$ \\
 & 8 & \boldmath{$113.24\pm0.08$} & $113.75\pm0.16$ & $113.60\pm0.15$ & $114.03\pm0.05$ & $113.59\pm0.06$ \\ \bottomrule
\end{tabular}
\end{table}




\subsection{Neural Architecture Search}
\label{section:nas}
The problem of NAS is finding the single best architecture in a very large space of architectures. Theoretically, this problem looks like
\begin{gather}
\label{eq:theoretical_nas}
    \min\limits_{a \in \mathcal{A}}\;\ell(a, \theta^*(a)) \quad \text{ where } \quad \theta^*(a) = \argmin\limits_{\theta \in \mathbb{R}^P} \;  \ell(a, \theta)
\end{gather}
where $\mathcal{A}$ is a discrete set of architectures, $a$ is a single architecture drawn from $\mathcal{A}$, and $\theta(a)$ represents the parameters for architecture $a$ (in our case the weights of a neural network). 

This is a difficult problem because of the properties of $\mathcal{A}$.
The size of $\mathcal{A}$ scales exponentially with every degree of freedom (e.g., adding a new operation).  
Set $\mathcal{A}$ is a discrete set as architectures consist of discrete components, so we cannot directly use tools from continuous optimization. 
Finally, each $\theta(a)$ is quite expensive to optimize in terms of computational cost and memory. 

 Two key relaxations make solving \eqref{eq:theoretical_nas} tractable. 
 The first utilizes weight sharing by training architectures together in a supernetwork \cite{pmlr-v80-pham18a}.
A supernetwork is a directed acyclic graph (DAG) where edges represent operations and nodes represent intermediate representations.
The second approximates $\theta^*$ with one step of gradient descent \cite{liu2018darts}. 
 With these two relaxations, the problem to solve is now
\begin{gather}
    \min\limits_{\pi \in \mathbb{R}^{n_e \times n_{op}}} \mathbb{E}_{a \sim p(\pi)}[\ell(a, \hat{\theta}_{t+1})] \quad \text{where} \quad \hat{\theta}_{t+1}=\theta_{t}-\eta \nabla_{\theta}\ell(a_t, \theta_t)
\end{gather}
We are learning a distribution $p(\pi)$ over $n_{op}^{n_e}$ the number of operations to the power of the number of edges in the DAG. 
The optimal distribution is the one that puts all the weight on the parameters $\pi$ which represent $a^*$ in the DAG. 

Most techniques in the NAS field either use a continuous distribution for $p(\pi)$ or use a deterministic parameterization with no sampling.
The reason for these choices is for ease of optimization. By using a continuous distribution they can take gradients by using either pathwise derivatives \cite{chen2021drnas} or the reparameterization trick \cite{xie2018snas}. By using a deterministic parameterization \cite{liu2018darts} \cite{Xu2020PC-DARTS:} they can directly compute the gradient. 
The authors of \cite{chen2021drnas} even note that discrete parameterizations suffer "instability".

We argue that it is better to use a discrete distribution to stay closer to the original optimization problem. By searching in the relaxed space significant bias is introduced to the problem and the final architecture requires a projection back to the discrete space.
Searching in the relaxed space also requires holding the entire supernetwork in memory.
Using a discrete distribution means searching directly in $\mathcal{A}$ and only having to hold a single path in memory so that discrete techniques can search directly on large-scale problems \cite{cai2018proxylessnas}.

The main problem with discrete distributions is estimating the gradient. We can use DBsurf to reduce variance and show that it is competitive with other highly engineered NAS techniques on NasBench201 \cite{Dong2020NAS-Bench-201:}, a common benchmark for NAS algorithms. 
For details about the search space and training procedure see Appendix \ref{appendix:nas}.

DBsurf is highly competitive with the current best NAS techniques as seen in Table \ref{table:2} and significantly improves over other discrete parameterizations. 
Of the high performing techniques, the Few Shot methods require training multiple supernetworks at the same time, and DrNAS must use a progressive learning scheme in order to search over larger problems.
DBsurf requires only single paths sampled from a supernetwork to be held in memory and, therefore, no special schemes to search on larger problems.
By tackling the difficulties of gradient estimation over discrete distributions, we can enjoy the benefits of reduced computational overhead, opportunities for parallelism, and no final projection, which come from searching in a discrete space. 
\begin{table}[h]
\small
\centering
\caption{The mean $\pm$ standard error over 4 seeds of the test accuracy of the selected architecture. The methods achieving the highest average test accuracy are \textbf{bolded}. The best performing architectures in the search space are denoted by \textbf{Optimal}. }
\label{table:2}
\begin{tabular}{@{}llccc@{}}
\toprule
\textbf{Parameterization}      & \multicolumn{1}{c}{\textbf{Method}} & \textbf{CIFAR-10}         & \textbf{CIFAR100}         & \textbf{ImgNet-16-120}    \\ \midrule
\multirow{2}{*}{Deterministic} & \textbf{DARTS} \cite{liu2018darts}                     & $54.30 \pm 0.00$          & $38.97 \pm 0.00$          & $18.41 \pm 0.00$          \\
                               & \textbf{PC-DARTS} \cite{Xu2020PC-DARTS:}                  & $93.41 \pm 0.00$          & $67.48 \pm 0.00$          & $41.32 \pm 0.22$          \\
                               &                                     & \multicolumn{1}{l}{}      & \multicolumn{1}{l}{}      & \multicolumn{1}{l}{}      \\
\multirow{4}{*}{Continuous}    & \textbf{SNAS} \cite{xie2018snas}                      & $92.77 \pm 0.83$          & $69.34 \pm 1.98$          & $43.16 \pm 2.64$          \\
                               & \textbf{Few Shot} \cite{pmlr-v139-zhao21d}                  & $93.88 \pm 0.25$          & $71.49 \pm 1.41$          & $\mathbf{46.43 \pm 0.19}$       \\
                               & \textbf{Few Shot GM} \cite{hu2022generalizing}               & $\mathbf{94.36 \pm 0.00}$ & $\mathbf{73.51 \pm 0.00}$ & $46.34 \pm 0.00$          \\
                               & \textbf{DrNAS} \cite{chen2021drnas}                     & $\mathbf{94.36 \pm 0.00}$ & $\mathbf{73.51 \pm 0.00}$ & $46.34 \pm 0.00$          \\
                               &                                     & \multicolumn{1}{l}{}      & \multicolumn{1}{l}{}      & \multicolumn{1}{l}{}      \\
\multirow{2}{*}{Discrete}      & \textbf{DSNAS} \cite{2020dsnas}                     & $93.08 \pm 0.29$          & $31.01 \pm 16.38$         & $41.31 \pm 0.22$          \\
                            & \textbf{GDAS} \cite{2019gdas}                     & $41.02 \pm 0.00$          & $24.20 \pm 8.08$         & $41.31 \pm 0.22$          \\
                               & \textbf{DBsurf}               & $\mathbf{94.36 \pm 0.00}$ & $\mathbf{73.51 \pm 0.00}$ & $46.34 \pm 0.00$ \\
                               & \textbf{Optimal}                    & \textbf{94.37}            & \textbf{73.51}            & \textbf{47.31}            \\ \bottomrule
\end{tabular}
\end{table}

\section{Conclusion}
Estimating the gradient of an expectation \(\nabla_\theta \mathbb{E}_{p_\theta}[f(x)]\) over a discrete distribution is a challenging task. 
This paper proposes a new stochastic estimator based on a novel sampling procedure that corrects inconsistencies between the target distributions and the drawn samples, a common issue in low-sampling regimes. 
From the methodological point of view, we develop a promising  low discrepancy sampling technique that is designed for discrete distributions and low sampling regimes. 
Such a sampling method moves beyond traditional Quasi Monte Carlo \cite{niederreiter1992random} approaches, which usually concentrate on continuous distributions and demand a large number of samples. 
The experiments show that this new technique leads to excellent performance in multiple very diverse problems, showing a lot of potential for further research. 
Possible extensions to the proposed method are tuning the correction parameter over training as in \cite{Titsias2021DoubleCV} or applying the sampling procedure to other estimators that use more sophisticated variance reduction strategies than LOORF. Finally, another promising line of work is the construction of new methods for Neural Architecture Search based on stochastic gradient estimators for discrete distributions. This would result in a significant reduction in the computational complexity compared to most of the current techniques, removing one of the main barriers to the further applicability of NAS.

\newpage

\bibliographystyle{plain}
\bibliography{literature}

\begin{thebibliography}{10}

\bibitem{Asmussen2007StochasticS}
S{\o}ren Asmussen and Peter~W. Glynn.
\newblock Stochastic simulation - algorithms and analysis.
\newblock In {\em Stochastic modeling and applied probability}, 2007.

\bibitem{Blackwell1947ConditionalEA}
David Blackwell.
\newblock Conditional expectation and unbiased sequential estimation.
\newblock {\em Annals of Mathematical Statistics}, 18:105--110, 1947.

\bibitem{cai2018proxylessnas}
Han Cai, Ligeng Zhu, and Song Han.
\newblock Proxyless{NAS}: Direct neural architecture search on target task and
  hardware.
\newblock In {\em International Conference on Learning Representations}, 2019.

\bibitem{chen2021drnas}
Xiangning Chen, Ruochen Wang, Minhao Cheng, Xiaocheng Tang, and Cho-Jui Hsieh.
\newblock Dr{\{}nas{\}}: Dirichlet neural architecture search.
\newblock In {\em International Conference on Learning Representations}, 2021.

\bibitem{Dimitriev2021ARMSAG}
Aleksandar Dimitriev and Mingyuan Zhou.
\newblock Arms: Antithetic-reinforce-multi-sample gradient for binary
  variables.
\newblock In {\em International Conference on Machine Learning}, 2021.

\bibitem{2019gdas}
Xuanyi Dong and Yi~Yang.
\newblock Searching for a robust neural architecture in four gpu hours.
\newblock {\em 2019 IEEE/CVF Conference on Computer Vision and Pattern
  Recognition (CVPR)}, Jun 2019.

\bibitem{Dong2020NAS-Bench-201:}
Xuanyi Dong and Yi~Yang.
\newblock Nas-bench-201: Extending the scope of reproducible neural
  architecture search.
\newblock In {\em International Conference on Learning Representations}, 2020.

\bibitem{disarm}
Zhe Dong, Andriy Mnih, and George Tucker.
\newblock Disarm: An antithetic gradient estimator for binary latent variables.
\newblock {\em CoRR}, abs/2006.10680, 2020.

\bibitem{figurnov2018implicit}
Mikhail Figurnov, Shakir Mohamed, and Andriy Mnih.
\newblock Implicit reparameterization gradients.
\newblock {\em Advances in neural information processing systems}, 31, 2018.

\bibitem{HOERNIG2018469}
Steffen Hoernig.
\newblock On the minimum correlation between symmetrically distributed random
  variables.
\newblock {\em Operations Research Letters}, 46(4):469--471, 2018.

\bibitem{hu2022generalizing}
Shoukang Hu, Ruochen Wang, Lanqing HONG, Zhenguo Li, Cho-Jui Hsieh, and Jiashi
  Feng.
\newblock Generalizing few-shot {NAS} with gradient matching.
\newblock In {\em International Conference on Learning Representations}, 2022.

\bibitem{2020dsnas}
Shoukang Hu, Sirui Xie, Hehui Zheng, Chunxiao Liu, Jianping Shi, Xunying Liu,
  and Dahua Lin.
\newblock Dsnas: Direct neural architecture search without parameter
  retraining.
\newblock {\em 2020 IEEE/CVF Conference on Computer Vision and Pattern
  Recognition (CVPR)}, Jun 2020.

\bibitem{jang2017categorical}
Eric Jang, Shixiang Gu, and Ben Poole.
\newblock Categorical reparameterization with gumbel-softmax.
\newblock In {\em International Conference on Learning Representations}, 2017.

\bibitem{Kingma2013AutoEncodingVB}
Diederik~P. Kingma and Max Welling.
\newblock Auto-encoding variational bayes.
\newblock {\em CoRR}, abs/1312.6114, 2013.

\bibitem{Kool2019Buy4R}
Wouter Kool, Herke van Hoof, and Max Welling.
\newblock Buy 4 reinforce samples, get a baseline for free!
\newblock In {\em DeepRLStructPred@ICLR}, 2019.

\bibitem{Kool2020EstimatingGF}
Wouter Kool, Herke van Hoof, and Max Welling.
\newblock Estimating gradients for discrete random variables by sampling
  without replacement.
\newblock {\em ArXiv}, abs/2002.06043, 2020.

\bibitem{omniglot}
Brenden~M. Lake, Ruslan Salakhutdinov, and Joshua~B. Tenenbaum.
\newblock Human-level concept learning through probabilistic program induction.
\newblock {\em Science}, 350(6266):1332--1338, 2015.

\bibitem{mnist}
Y.~Lecun, L.~Bottou, Y.~Bengio, and P.~Haffner.
\newblock Gradient-based learning applied to document recognition.
\newblock {\em Proceedings of the IEEE}, 86(11):2278--2324, 1998.

\bibitem{liu2018darts}
Hanxiao Liu, Karen Simonyan, and Yiming Yang.
\newblock {DARTS}: Differentiable architecture search.
\newblock In {\em International Conference on Learning Representations}, 2019.

\bibitem{leakyrelu}
A.~L. Maas, A.~Y. Hannun, and A.~Y. Ng.
\newblock Rectifier nonlinearities improve neural network acoustic models.
\newblock In {\em In ICML Workshop on Deep Learning for Audio, Speech and
  Language}, 2013.

\bibitem{maddison2017the}
Chris~J. Maddison, Andriy Mnih, and Yee~Whye Teh.
\newblock The concrete distribution: A continuous relaxation of discrete random
  variables.
\newblock In {\em International Conference on Learning Representations}, 2017.

\bibitem{JMLR:v21:19-346}
Shakir Mohamed, Mihaela Rosca, Michael Figurnov, and Andriy Mnih.
\newblock Monte carlo gradient estimation in machine learning.
\newblock {\em Journal of Machine Learning Research}, 21(132):1--62, 2020.

\bibitem{niederreiter1992random}
Harald Niederreiter.
\newblock {\em Random number generation and quasi-Monte Carlo methods}.
\newblock SIAM, 1992.

\bibitem{owenbook}
Art~B. Owen.
\newblock {\em Monte Carlo theory, methods and examples}.
\newblock 2013.

\bibitem{NEURIPS2019_9015}
Adam Paszke, Sam Gross, Francisco Massa, Adam Lerer, James Bradbury, Gregory
  Chanan, Trevor Killeen, Zeming Lin, Natalia Gimelshein, Luca Antiga, Alban
  Desmaison, Andreas Kopf, Edward Yang, Zachary DeVito, Martin Raison, Alykhan
  Tejani, Sasank Chilamkurthy, Benoit Steiner, Lu~Fang, Junjie Bai, and Soumith
  Chintala.
\newblock Pytorch: An imperative style, high-performance deep learning library.
\newblock In {\em Advances in Neural Information Processing Systems 32}, pages
  8024--8035. Curran Associates, Inc., 2019.

\bibitem{pmlr-v80-pham18a}
Hieu Pham, Melody Guan, Barret Zoph, Quoc Le, and Jeff Dean.
\newblock Efficient neural architecture search via parameters sharing.
\newblock In Jennifer Dy and Andreas Krause, editors, {\em Proceedings of the
  35th International Conference on Machine Learning}, volume~80 of {\em
  Proceedings of Machine Learning Research}, pages 4095--4104. PMLR, 10--15 Jul
  2018.

\bibitem{rao1945information}
C~Radhakrishna Rao.
\newblock Information and the accuracy attainable in the estimation of
  statistical parameters.
\newblock {\em Reson. J. Sci. Educ}, 20:78--90, 1945.

\bibitem{Robert2005MonteCS}
Christian~P. Robert and George Casella.
\newblock Monte carlo statistical methods.
\newblock {\em Technometrics}, 47:243 -- 243, 2005.

\bibitem{Ruiz2016TheGR}
Francisco J.~R. Ruiz, Michalis~K. Titsias, and David~M. Blei.
\newblock The generalized reparameterization gradient.
\newblock In {\em NIPS}, 2016.

\bibitem{Shi2022GradientEW}
Jiaxin Shi, Yuhao Zhou, Jessica Hwang, Michalis~K. Titsias, and Lester~W.
  Mackey.
\newblock Gradient estimation with discrete stein operators.
\newblock {\em ArXiv}, abs/2202.09497, 2022.

\bibitem{Sutton1998}
Richard~S. Sutton and Andrew~G. Barto.
\newblock {\em Reinforcement Learning: An Introduction}.
\newblock The MIT Press, second edition, 2018.

\bibitem{Titsias2021DoubleCV}
Michalis~K. Titsias and Jiaxin Shi.
\newblock Double control variates for gradient estimation in discrete latent
  variable models.
\newblock In {\em International Conference on Artificial Intelligence and
  Statistics}, 2021.

\bibitem{NIPS2015_1373b284}
Michalis Titsias RC~AUEB and Miguel L\'{a}zaro-Gredilla.
\newblock Local expectation gradients for black box variational inference.
\newblock In C.~Cortes, N.~Lawrence, D.~Lee, M.~Sugiyama, and R.~Garnett,
  editors, {\em Advances in Neural Information Processing Systems}, volume~28.
  Curran Associates, Inc., 2015.

\bibitem{Tucker2017REBARLU}
G.~Tucker, Andriy Mnih, Chris~J. Maddison, John Lawson, and Jascha~Narain
  Sohl-Dickstein.
\newblock Rebar: Low-variance, unbiased gradient estimates for discrete latent
  variable models.
\newblock In {\em NIPS}, 2017.

\bibitem{tucker2017rebar}
George Tucker, Andriy Mnih, Chris~J. Maddison, Dieterich Lawson, and Jascha
  Sohl-Dickstein.
\newblock Rebar: Low-variance, unbiased gradient estimates for discrete latent
  variable models, 2017.

\bibitem{Williams1992SimpleSG}
Ronald~J. Williams.
\newblock Simple statistical gradient-following algorithms for connectionist
  reinforcement learning.
\newblock {\em Machine Learning}, 8:229--256, 1992.

\bibitem{fashionmnist}
Han Xiao, Kashif Rasul, and Roland Vollgraf.
\newblock Fashion-mnist: a novel image dataset for benchmarking machine
  learning algorithms.
\newblock {\em CoRR}, abs/1708.07747, 2017.

\bibitem{xie2018snas}
Sirui Xie, Hehui Zheng, Chunxiao Liu, and Liang Lin.
\newblock {SNAS}: stochastic neural architecture search.
\newblock In {\em International Conference on Learning Representations}, 2019.

\bibitem{Xu2020PC-DARTS:}
Yuhui Xu, Lingxi Xie, Xiaopeng Zhang, Xin Chen, Guo-Jun Qi, Qi~Tian, and
  Hongkai Xiong.
\newblock Pc-darts: Partial channel connections for memory-efficient
  architecture search.
\newblock In {\em International Conference on Learning Representations}, 2020.

\bibitem{Yin2018ARMAG}
Mingzhang Yin and Mingyuan Zhou.
\newblock Arm: Augment-reinforce-merge gradient for stochastic binary networks.
\newblock In {\em International Conference on Learning Representations}, 2018.

\bibitem{pmlr-v139-zhao21d}
Yiyang Zhao, Linnan Wang, Yuandong Tian, Rodrigo Fonseca, and Tian Guo.
\newblock Few-shot neural architecture search.
\newblock In Marina Meila and Tong Zhang, editors, {\em Proceedings of the 38th
  International Conference on Machine Learning}, volume 139 of {\em Proceedings
  of Machine Learning Research}, pages 12707--12718. PMLR, 18--24 Jul 2021.

\end{thebibliography}

\newpage
\appendix
\startcontents[sections]
\printcontents[sections]{l}{1}{\setcounter{tocdepth}{2}}

\section{Theory}
\label{appendix:theory}
\subsection{Eq. \ref{eq:6}:  LOORF is Reinforce with a discrepancy correction term}

\begin{proof}

\begin{align}
 g_{\scalebox{0.5}{LOORF}}(\textbf{x}) &= \frac{1}{n}\sum_{i=1}^n \Bigl(f(x_i)-\frac{1}{n-1}\sum_{j\neq i}f(x_j)\Bigr) \nabla_\theta \log p(x_i) \label{eq:loorf1}\\
    &= \frac{1}{n-1}\sum_{i=1}^n \Bigl(f(x_i)-\frac{1}{n}\sum_{j=1}^n f(x_j)\Bigr) (x_i - p) \label{eq:loorf2}\\
    &= \frac{1}{n-1}\sum_{i=1}^n f(x_i)(x_i - p) - \frac{1}{n-1}\sum_{i=1}^n (x_i - p) \frac{1}{n}\sum_{j=1}^n f(x_j) \label{eq:loorf3}\\
    &= \frac{1}{n-1}\sum_{i=1}^n f(x_i)(x_i - p) - (\frac{1}{n}\sum_{i=1}^nx_i - p) \frac{1}{n-1}\sum_{j=1}^n f(x_j) \label{eq:loorf4}  \\
 &= \underbrace{\frac{1}{n-1}\sum_{i=1}^{n}f(x_i)(x_i-p)}_{\text{Reinforce}} - \underbrace{(\hat{p}-p)\frac{1}{n-1}\sum_{i=1}^{n}f(x_i)}_{\text{Discrepancy correction}}
\end{align}
In \eqref{eq:loorf2} we use $\nabla_\theta \log p(x_i) = (x_i - p)$ and re-arrange terms by splitting $f(x_i) = \frac{n}{n-1}f(x_i) - \frac{1}{n-1}f(x_i)$ to remove the $j \neq i$ term from the second sum. Then, in \eqref{eq:loorf3}, \eqref{eq:loorf4} we simply re-arrange terms using linearity.
\end{proof}

\subsection{Proposition 2.2}
\begin{namedthm*}{Proposition 2.2} For any \(p \in [0, 1]^d\) and any \(n\in \mathbb{N}\) if we take $
\alpha \leq \min \left( \frac{1-p_i}{p_i}, \frac{p_i}{1-p_i}\right)
$ for any \(i\), the sampling procedure in \hyperref[alg:algorithm]{Algorithm 1} preserves the original distribution.  Equivalently: $$\E[x_k] = p \quad  k=1,...,n$$
\begin{proof}
We denote by $q_1,...,q_n$ the different parameters generated by the sampling algorithm and $x_1,...,x_n$ the associated random variables, using $q_k^{i}$ to refer to the \(i\text{-th}\) component of the parameter in the \(k\text{-th}\) step of the algorithm.
First we can show that for the choices of $\alpha$ the sampling algorithm always takes $$q_k = p(1+\alpha) -\alpha\hat{p}_{k-1} \qquad k=2,...,n .$$ 
This can be seen bounding $q_k$ by taking into account that $\hat{p}_{k-1}\in [0, 1]^d$. For an arbitrary \(i \in \{1,...,d\}\) we have:
$$
p_i - \alpha (1-p_i) \leq q_k^{i} \leq p_i(1+\alpha)
$$
Note that the left hand side is non negative by the fact that $\alpha \leq \frac{p_i}{1-p_i}$ and the right hand side is bounded by 1 from $\alpha \leq \frac{1-p_i}{p_i}$. Therefore there is no clamping in the sampling algorithm, i.e \(q_k = p(1+\alpha) -\alpha\hat{p}_{k-1} \; \forall k\).

Next, we claim that $\E[\hat{p}_k] = p \;$ for any $k$ when the statistic $\hat{p}_k$ consists of the samples generated by $q_1,...,q_k$. If the claim holds the result is direct
$$
\E[x_k] = \E[\E[x_k \mid \hat{p}_{k-1}]] = (1+\alpha)p - \alpha \E[\hat{p}_{k-1}] = p 
$$
using the Tower Law in the first identity and that $q_k = p(1+\alpha) -\alpha\hat{p}$ (as just proven) in the second identity.

Finally, to proof the claim we can proceed by induction. The base case follows by construction of the sampling procedure, since we defined $q_1 = p$. Now assume that the claim holds for $k-1$ and let's prove it for $\hat{p}_k$.
$$
\E[\hat{p}_k] = \E[\:\E[\hat{p}_k \mid \hat{p}_{k-1}]\:] = \frac{k-1}{k}\E[\hat{p}_{k-1}] + \frac{1}{k} \E[\:\E[x_k\mid \hat{p}_{k-1}]\:] = \frac{k-1}{k}p + \frac{1}{k}p(1+\alpha) -\frac{\alpha}{k}p = p
$$
\end{proof}
\end{namedthm*}

\subsection{Proposition 2.1}
\begin{namedthm*}{Proposition 2.1}
Let $\textbf{x}=\{x_1,...,x_n\}$ be a set of i.i.d random variables with distribution $Bernoulli(p)$ for arbitrary $p \in (0, 1)^d$ and $n \geq 2$. Consider the set  $\textbf{z}=\{z_1,...,z_n\}$ of random variables associated to Algorithm 1 with initial parameters $p, n$ and $0 < \alpha \leq \min \left( \frac{1-p_i}{p_i}, \frac{p_i}{1-p_i}\right)$ for any \(i \in \{1,...,d\}\). We have:
$$
\mathbb{E}[\|D_p(\textbf{z})\|^2] < \mathbb{E}[\|D_p(\textbf{x})\|^2]
$$
\begin{proof}
    We know from the previous lemma that when taking $\alpha \leq \min \left( \frac{1-p_i}{p_i}, \frac{p_i}{1-p_i}\right)$ DBsample   preserves the marginal distributions of the $z_k$. Then, it follows
    $$
    \mathbb{E}[\|D_p(\textbf{z})\|^2] = \mathbb{E}[\|\hat{p}_n-p\|^2]= \E[\langle \frac{1}{n}\sum_{k=1}^n z_k-p, \frac{1}{n}\sum_{k=1}^n z_k-p \rangle] = \frac{1}{n^2}\sum_{k, \ell} \E[ \langle z_k, z_\ell \rangle]-\|p\|^2 
    $$
    $$
     = \frac{1}{n^2}\sum_{k,  \ell}\sum_{r=1}^d Cov(z_k^r, z_\ell^r)
     = \frac{1}{n^2}\sum_{k\neq \ell}\sum_{r=1}^d Cov(z_k^r, z_\ell^r) + \frac{1}{n}\langle p, \mathbf{1_d}-p\rangle 
    $$
    where $\mathbf{1_d}$ denotes a d-dimensional vector of ones and \(z_k^r\) the \(r\text{-th}\) component of the random variable associated to the \(k\text{-th}\) step. Note that the previous argument holds for a general set of samples $\textbf{z}$ as far as the marginal distribution is preserved, $\mathbb{E}[z_k]=p \; \forall k$ since that is the only fact we used about $\textbf{z}$. Thus, in particular we have that for the i.i.d samples $x_1,...,x_n$
    $$
    \mathbb{E}[\|D_p(\textbf{x})\|^2] = \frac{1}{n}\langle p, \mathbf{1_d}-p\rangle 
    $$
    Consequently, to proof that $
\mathbb{E}[\|D_p(\textbf{z})\|^2] < \mathbb{E}[\|D_p(\textbf{x})\|^2]
$ it suffices to show that $Cov(z_k^r, z_\ell^r)\leq 0$ for any $k\neq \ell$ and any $r$ with strict inequality when $k=1, \ell=2$. 

We start showing that $Cov(z_1^r, z_2^r) < 0$ for any $r \in \{1,...,d\}$. From the proof of the previous lemma we know that the assumption on $\alpha$ implies that the sampling algorithm always generates $q_k = p(1+\alpha) -\alpha\hat{p}_{k-1}$. It follows that
$$
Cov(z_1^r, z_2^r) = p_r(\mathbb{E}[z_2^r=1 \mid z_1^r=1]-p_r)=-\alpha p_r(1-p_r) < 0 \qquad \forall r
$$
Next, we want to show that $Cov(z_k^r, z_\ell^r)\leq 0$ for any $k\neq \ell$ and any $r$. Assume, without loss of generality that $k < \ell$ and fix an arbitrary $r\in \{1,...,d\}$. Similarly as before we have
\begin{align*}
Cov(z_k^r, z_\ell^r) &= p_r(\mathbb{E}[z_\ell^r=1 \mid z_k^r=1]-p_r) = p_r(\mathbb{E}[\mathbb{E}[z_\ell^r=1 \mid z_k^r=1, \hat{p}_{\ell-1}^r] \mid z_k^r=1]-p_r)    \\
 &=  p_r(\alpha p_r - \alpha \mathbb{E}[\hat{p}_{\ell-1}^r \mid z_k^r=1]) = \alpha p_r( p_r -  \frac{1}{\ell-1}\sum_{j=1}^{\ell-1}\mathbb{E}[z_j^r=1 \mid z_k^r=1])  \\
 &= -\frac{\alpha}{\ell-1}( p_r(1-p_r) + \sum_{\substack{j < \ell \\  j\neq k}}Cov(z_k^r, z_j^r))  
\end{align*}
From the last line we see that showing that $Cov(z_k^r, z_\ell^r)\leq 0$  is equivalent to show that \[S_r(k, \ell-1) \geq -p_r(1-p_r)\] where we define \(S_r\) as:
$$S_r(k, \ell) = \sum_{\substack{j\leq \ell \\ j \neq k}}Cov(z_k^r, z_j^r) $$  
For \(k\geq 1, \ell  > k\), we have the recursive relation:
$$
S_r(k, \ell) = S_r(k, \ell-1) -\frac{\alpha}{\ell-1}(p_r(1-p_r)+S_r(k, \ell-1)) 
$$
From the fact that $\alpha \leq 1$ we have that  $S(1, \ell)$ is non decreasing in $S(1, \ell-1)$. Combining that with  $S(1, 2)\geq -p_r(1-p_r)$ we get that $S(1, \ell)\geq -p_r(1-p_r)$ for any $\ell$. Note that the later implies
$$Cov(z_1^r, z_\ell^r)= S(1, \ell)-S(1, \ell-1)= -\frac{\alpha}{j-1}(p_r(1-p_r)+S(1, \ell-1))\leq 0 \qquad \forall \ell \geq 2$$
From here we can proof by induction that $Cov(z_k^r, z_\ell^r)\leq 0 \quad \forall k, \ell \quad k \neq \ell$. \newline We proceed by strong induction on \(k\). We just proved the base case \(k=1\), so let's proceed to the induction step. Suppose that:
$$
Cov(z_{j}^r, z_\ell^r)\leq 0 \quad \forall j \; : \; j < k \quad  \forall \ell \; : \; \ell \neq j
$$
we want to show that $Cov(z_{k}^r, z_\ell^r)\leq 0 \quad \forall \ell \; : \; \ell \neq k$. Note, that using the induction hypothesis we already have $Cov(z_{k}^r, z_\ell^r)\leq 0 \quad \forall \ell \; : \; \ell < k$. Also, for any \(\ell < k\) we have \(S_r(\ell, k-1) \leq 0\) and consequently:
\begin{align*}
Cov(z_{k}^r, z_\ell^r) &= -\frac{\alpha}{k-1}(p_r(1-p_r) + S_r(\ell, k-1))   \geq -\frac{\alpha}{k-1}p_r(1-p_r)
\end{align*}
Then, it follows:
\[
S_r(k, k) = \sum_{\ell < k} Cov(z_{k}^r, z_\ell^r) \geq -p_r(1-p_r) 
\]
Now, we can use  the recursive relation on $S_r(k, \ell)$ for \(l > k\). Just as before, from the fact that \(\alpha \leq 1\) we have that $S_r(k, \ell)$ is increasing in $S_r(k, \ell-1)$. Combining that with $S_r(k, k) \geq -p_r(1-p_r)$ we get that $S_r(k, \ell) \geq -p_r(1-p_r)$ for any \(\ell \geq k\). Finally this implies
$$Cov(z_k^r, z_\ell^r)= S_r(k, \ell)-S_r(k, \ell-1)= -\frac{\alpha}{j-1}(p_r(1-p_r)+S_r(k, \ell-1))\leq 0 \qquad \forall \ell > k$$
Therefore we can conclude that $Cov(z_k^r, z_\ell^r)\leq 0 \quad \forall k, \ell \quad k \neq \ell$. Since $r$ was chosen to be arbitrary, it holds for any $r$. This combined with $Cov(z_1^r, z_2^r) < 0$ for any $r \in \{1,...,d\}$ (previously proven) allows us to conclude:
$$
\mathbb{E}[\|D_p(\textbf{z})\|^2] < \mathbb{E}[\|D_p(\textbf{x})\|^2]
$$
\end{proof}
\end{namedthm*}

\subsection{Eq. \ref{eq:var_g}: Computation for $\V[g_q(x_1, x_2)]$}
On the one hand, the exact gradient is:
\begin{gather}
    \label{eq:exact_gradient}
    \nabla_\theta \mathbb{E}_{p_\theta}[f(x)] = \mathbb{E}_{p_\theta}[f(x)\nabla_\theta \log p_\theta(x)] = \mathbb{E}[f(x)(x-p)] = (f(1) - f(0))p(1-p)
\end{gather}
where \(p = p_\theta(x=1)\). Since the gradient only depends on \(\theta\) through Bernoulli parameter \(p\) we can simply identify \(p_\theta\) by p. \newline
Consider the general class of estimators of the form
\begin{equation}
    g_q(x_1, x_2) = \frac{p(1-p)}{\kappa_q}(f(x_1)-f(x_2))(x_1-x_2)  \qquad (x_1, x_2) \sim q
\end{equation}
with \(q\) probability distribution in \(\{0, 1\}^2\) and \(\kappa_q\) a constand depending on \(q\). The expected value is
\[
\mathbb{E}_q[g_q(x_1,x_2)] = \frac{p(1-p)}{\kappa_q}(f(1)-f(0))(q(x_1=1, x_2=0)+q(x_1=0. x_2=0))  
\]
so if we want it to be unbiased we must simply take \(\kappa_q = q(x_1=1, x_2=0)+q(x_1=0, x_2=0)\). With that choice, the variance is:
\[
\mathbb{V}ar_q[g_q(x_1, x_2)] = \mathbb{E}_q[g_q(x_1,x_2)^2]-\mathbb{E}_q[g_q(x_1,x_2)]^2 = [p(1-p)(f(1)-f(0))]^2\frac{1-\kappa_q}{\kappa_q}
\]
We want to see how the correction parameter \(\alpha\) in our algorithm relates to variance. For that we simply have to look at the previously computed variance when taking \(q\) as the distribution associated to our sampling algorithm for \(d=1, n=2\) and a certain choice of \(\alpha\).
\begin{itemize}
    \item \(\alpha = 0\)
    \[
    \kappa_q = 2p(1-p)
    \]
    \[
    \mathbb{V}ar_q[g_q(x_1, x_2)] =  p(1-p)(f(1)-f(0))^2 \left(\frac{1}{2}-p(1-p)\right)
    \]
    \item \(\alpha \leq \min \left(\frac{p}{1-p}, \frac{1-p}{p} \right)\)
    \[
    \kappa_q = 2\mathbb{P}(x_1=0, x_2=1)=2(1-p)\mathbb{P}(x_2=1\mid x_1=0)=2p(1-p)(1+\alpha)
    \]
    where we used that for these values of \(\alpha\) there is no clamping and the marginal distributions are preserved (proved in  \hyperref[lemma:2]{Proposition 2}). 
    \[
    \mathbb{V}ar_q[g_q(x_1, x_2)]  = p(1-p)(f(1)-f(0))^2 \left(\frac{1}{2(1+\alpha)}-p(1-p)\right)
    \]
    \item \( \min \left(\frac{p}{1-p}, \frac{1-p}{p} \right) < \alpha \leq \max \left(\frac{p}{1-p}, \frac{1-p}{p} \right)\)\newline 

    If \(p\geq \frac{1}{2}\):
    \[
    \mathbb{P}(x_1=1,x_2=0)=p\mathbb{P}(x_2=0\mid x_1=1)=p(1-p)(1+\alpha)
    \]
    \[
    \mathbb{P}(x_1=0, x_2=1)=(1-p)\mathbb{P}(x_2=1\mid x_1=0)=1-p
    \]
    Hence:
    \[
    \kappa_q = p(1-p)(1+\alpha) + 1-p
    \]
    By symmetry if \(p < \frac{1}{2}\) we get \(\kappa_q = p(1-p)(1+\alpha) + p\) so in general we can write:
    \[
    \kappa_q = p(1-p)(1+\alpha) + \min(p, 1-p) = p(1-p) \left(1+\alpha + \frac{1}{\max(p, 1-p)}\right)
    \]
     \[
    \mathbb{V}ar_q[g_q(x_1, x_2)]  = p(1-p)(f(1)-f(0))^2 \left(\frac{\max(p, 1-p)}{(1+\alpha)\max(p, 1-p) + 1}-p(1-p)\right)
    \]
    \item \(\alpha > \max \left(\frac{p}{1-p}, \frac{1-p}{p} \right) \)
    \[
    \mathbb{P}(x_1=1,x_2=0)=p\mathbb{P}(x_2=0\mid x_1=1)=p
    \]
    \[
    \mathbb{P}(x_1=0, x_2=1)=(1-p)\mathbb{P}(x_2=1\mid x_1=0)=1-p
    \]
    \[
    \kappa_q=1 \quad \rightarrow \quad \mathbb{V}ar_q[g_q(x_1, x_2)]=0
    \]
\end{itemize}
\textbf{Remark: } In this example the variance is monotone decreasing in \(\alpha\). Indeed, note that within each interval the variance is monotone increasing in \(\frac{1}{1+\alpha}\) and that the different variance expressions given always match at the boundary of the intervals.
\begin{figure}[h]
    \centering
    \begin{subfigure}[b]{0.48\textwidth}
        \includegraphics[width=\textwidth]{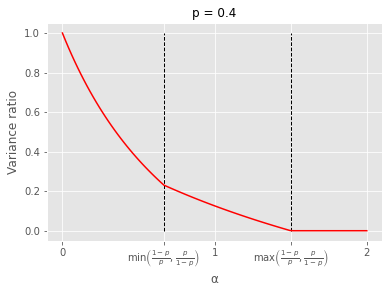}
    \end{subfigure}
    \hfill
    \begin{subfigure}[b]{0.48\textwidth}
 \includegraphics[width=\textwidth]{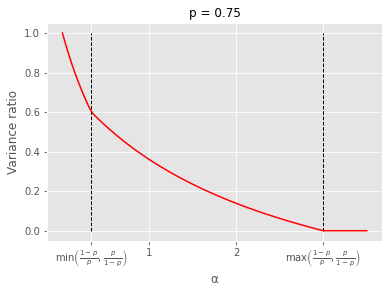}
    \end{subfigure}
    \caption{The ratio between the variance of DBsurf and the variance of LOORF is ploted across the different values of \(\alpha\).}
\end{figure}

\subsection{Fig. \ref{fig:corr}: Correlation Structure of DBshrink}

We denote $q_1,...,q_n$ the different parameters generated by the sampling algorithm and $x_1,...,x_n$ the associated random variables.  As a reminder the sampling procedure is defined as:
$$
q_n = \varphi (p(1+\alpha) - \alpha\hat{p}_{n-1}) \qquad \qquad q_1=p \qquad \qquad  \varphi(x) = \begin{cases}1 & x > 1 \\
0 & x < 0 \\
x & \text{otherwise}
\end{cases}
$$

Define $\nu_{n, k} = \P(\omega:\hat{p}_{n}(\omega) = \frac{k}{n})$ which trivially satisfies $\sum_{k=0}^{n}\nu_{n, k} = 1$.
We can compute the $\nu_{n, k}$ recursively by noting that:
$$
\nu_{n, k} = (1-\varphi(p(1+\alpha)-\frac{\alpha k}{n-1}))\nu_{n-1,k}
+ \varphi(p(1+\alpha)-\frac{\alpha (k-1)}{n-1})\nu_{n-1, k-1}
$$
with boundary conditions : $\nu_{n,k} = 0$ if $k > n$, $k < 0$  and $\nu_{1, 0} = 1-p, \; \nu_{1, 1}=p$.
The expression for $\nu_{n, k}$ comes from the law of total probability where we sum across the ways which the event $\hat{p}_{n} = \frac{k}{n}$ can occur.
$$\nu_{n, k} = \P \left(x_n = 0 \mid \hat{p}_{n-1}(w) = \frac{k}{n}\right) \P \left(\hat{p}_{n-1}(w) = \frac{k}{n}\right) + \P \left(x_n = 1 \mid \hat{p}_{n-1}(w) = \frac{k-1}{n}\right) \P \left(\hat{p}_{n-1}(w) = \frac{k-1}{n}\right) $$

To calculate the expected value of sample n, we can use the following recursive formulation
\begin{align}
\label{eq:expected_value}
    \E[x_n] &= \sum_{k=1}^{n}k\nu_{n,k} - \sum_{k=1}^{n-1}\E[x_k] \qquad \E[x_1]=p
\end{align}
When $\alpha > \text{min}(\frac{(1-p)}{p}, \frac{p}{1-p)})$ the marginal distribution for each sample changes, and this procedure gives us the expected value of the marginal distribution for each sample.

We are interested in studying the correlation between the two Bernoulli random variables involved in the estimator.
\begin{gather*}
    \text{Corr}(x_i, x_j) = \frac{\text{Cov}(x_i, x_j)}{\sqrt{\V{(x_i)}}\sqrt{\V{(x_j)}}} = \frac{\text{Cov}(x_i, x_j)}{\sqrt{\E[x_i](1-\E[x_i]))}\sqrt{\E[x_j](1-\E[x_j]))}}
\end{gather*}
We have a procedure to calculate the expected value of the marginal distribution \eqref{eq:expected_value}, so we are left to compute 
\begin{gather*}
    \text{Cov}(x_i, x_j) = \mathbb{E}[x_i x_j] - \E[x_i]\E[x_j] = \mathbb{P}(x_i = 1, x_j = 1) - \E[x_i]\E[x_j]
\end{gather*}
We directly compute $\mathbb{P}(x_i = 1, x_j = 1)$: 

Define $S = \{0, 1\}^j$ the set of all binary strings of length $j$ and $G = \{s \in S | s_i = 1, s_j = 1\}$ we then calculate
\begin{gather*}
    \mathbb{P}(x_i = 1, x_j = 1) = \sum_{g \in G} \P(x_1 = g_1, \dots, x_j = g_j) \\
    = \sum_{g \in G} \P(x_1 = g_1) \prod_{l=2}^j \P(x_l = g_l | x_1 = g_1, \dots, x_{l-1} = g_{l-1})
\end{gather*}
We can calculate this expression recursively by starting at the bottom of a binary tree and computing these probabilities at every branch. 
\newline

\subsection{Fig. \ref{fig:toy_variance}: Debiasing DBsurf in 1-D}

The LOORF estimator can be expressed as a product of differences \cite{Dimitriev2021ARMSAG}
$$
g_n = \frac{1}{n-1}\sum_{i=1}^{n}(f(x_i)-\hat{\mu})(x_i-p) = \frac{1}{n(n-1)} \sum_{i< j}(f(x_i)-f(x_j))(x_i-x_j)
$$
In the 1D case we can write the expectation of LOORF as
$$
\E[g_n] = \frac{1}{n(n-1)} \sum_{i < j}(f(1)-f(0))(\mathbb{P}(x_i=1, x_j=0) + \mathbb{P}(x_i=0, x_j=1))
$$

We define $S_{ij} = \mathbb{P}(x_j=1 | x_i=1)$ and $\mu_i = \E[x_i]$. With our sampling procedure, we have
$$
\mathbb{P}(x_i=1, x_j=0) = \mathbb{P}(x_i=1)-\mathbb{P}(x_i=1, x_j=1) = \mu_i(1-S_{ij})
$$
$$
\mathbb{P}(x_i=0, x_j=1) = \mathbb{P}(x_j=1)-\mathbb{P}(x_i=1, x_j=1) = \mu_j - \mu_iS_{ij}
$$
Then, in the one-dimensional case the expected value of the estimator is:
$$
\E[g_n] = \frac{1}{n(n-1)} \sum_{i < j}(f(1)-f(0))((1-S_{ij})\mu_i  + \mu_j - \mu_i S_{ij}) = \frac{1}{n(n-1)}(f(1)-f(0))\sum_{i < j}(\mu_i + \mu_j - 2\mu_i S_{ij})
$$
Recall that the true gradient from \eqref{eq:exact_gradient} is:
$$
g^* = (f(1)-f(0))p(1-p)
$$
so that we can debias our estimator by using a multiplicative debiasing term of
$$\frac{n(n-1)p(1-p)}{\sum_{i < j}(\mu_i + \mu_j -2\mu_iS_{ij})}$$

\section{General Discrete Distributions}
\label{appendix:general_discrete}
In this section of the appendix we expand on \ref{section:general_discrete} to describe how our sampling algorithm can be used with a wide range of discrete distributions. Indeed, the only constraint of our algorithm is that the distribution must be independent across the dimensions, due to the fact that our algorithm runs independently in each dimension. Thus, for clarity we describe the 1D case, but noting that the multidimensional case is just the replication of the 1D procedure individually in each dimension.
\subsection{Categorical distributions}
Consider a categorical distribution over \(m\) categories, identified by \(\{1,...,m\}\). The distribution is determined by a vector \(\textbf{p}=(p_1,...,p_m)\) such that \(\sum_{i=1}^{m}p_i=1\), inducing a probability mass function:
\[
\mathbb{P}_\textbf{p}(x=i)=p_i
\]
Note, that to each of the categories we can associate a random variable following a Bernoulli distribution with parameter \(p_i\), by just considering
\[
x^{(i)}(\omega) = 1_{\{x(\omega)=i\}}
\]
From this observation, we can apply the correction of the sampling algorithm  to each of the categories as if they were Bernoullis. Then we can create a modified categorical distribution by simply putting together each of the modified \(p_i\) with the proper normalization. More concretely:

        \begin{algorithm}[H]
            \caption{(Categorical DBsample)}
            \begin{algorithmic}
             
                \State \textbf{Input: } \(\textbf{p}\) (true probabilities), \(m\) (categories), \(n\) (samples), \(\alpha\) (correction factor) 
                \State \vspace{-5pt}
                \State $\hat{\textbf{p}}$ $\gets$ $\textbf{p}$
                \State $\textbf{q}$ $\gets$ $\textbf{p}$
                
                \For{\(k=1,...,n\)}
                    \State \(x_k\sim Categorical(\textbf{q})\)
                    \For{\(r=1,...,m\)}
                    \State $\hat{p}^{(r)} = ((k-1) \cdot \hat{p}^{(r)} + x_k^{(r)})/i$
                    \State \(q^{(r)} = \min(1, (p_r (1+\alpha) - \alpha \hat{p}^{(r)})_{+})\)
                \EndFor
                 \State \(q = \frac{q}{\|q\|_1} \)
            \EndFor
                \State \vspace{-5pt}
                \State \textbf{Output: }\(\{x_1,...,x_n\}\)
            \end{algorithmic}
        \end{algorithm}

\subsection{Discrete distributions with finite support}
In the case that we want to sample from a discrete probability distribution with finite support (e.g binomial, hypergeometric) we can simply interpret it as a categorical. Recall that any discrete distribution with finite support is characterized by a vector of probabilities \(\textbf{p}=(p_1,...,p_m) \) and a set of points \(\{a_1,...,a_m\}\). 
\[
\mathbb{P}_\textbf{p}(x=a_k)=p_k
\]
Therefore, it is clear that to sample from  such a distribution we can simply  use the transformation:
\[
\Tilde{x}(\omega) = \sum_{i=1}^{m}a_i1_{\{x(\omega)=i\}}
\]
so that it suffices to sample \(x\sim Categorical(\textbf{p})\).
\subsection{Discrete distributions with infinite support}
 Unfortunately the previous procedure does not directly apply here, since we cannot identify these distributions with categoricals. However we can group points in the tails in such a way that we reduce it to the finite support case. Consider a discrete distribution such that its support \(S\) is contained in \((b, +\infty)\) for some \(b \in \mathbb{R}\) and that \(S \cap [b, c]\) is finite for any \(c \in \mathbb{R}\). This includes the most important discrete distributions that have infinite support (Poisson, Negative Binomial, Geometric,...). In that case, for every \(\epsilon > 0 \) we have that that \(F^{-1}([0, 1-\epsilon))\) is finite so we can reduce the sampling to the categorical case again in the following way.
\begin{enumerate}
    \item Fix \(\epsilon > 0\) and consider the antiimage of the cdf: \(\{a_1,...,a_m\} = F^{-1}([0, 1-\epsilon))\) and its complementary (w.r.t the support) set \(U\),
    \item Set \(\textbf{p}=(p_1,...,p_{m}, p_{m+1})\) such that: 
    \[p_k=\mathbb{P}(x=a_k) \quad  k=1,...,m  \qquad  \; p_{m+1} = \mathbb{P}(x\in U)\]
   \item[]  \hspace{-12pt}(Repeat \(n\) times)
    \item Sample \(x \sim Categorical(\textbf{p})\) using Algorithm 2 and \(z \sim \mathbb{P}(\cdot \mid U)\) 
    \item Apply the transformation:
    \[
    \Tilde{x} = \sum_{i=1}^{m}a_i1_{\{x=i\}} + z1_{\{x=m+1\}}
    \]

\end{enumerate}
In practice \(z\) doesn't need to be sampled unless \(x=m+1\). Also note that in the (uncommon) case that the support is contained in a set of the form \((-\infty, c)\) we simply have to replace the right tail by the left tail in (1), that is \(\{a_1,...,a_m\} = F^{-1}((\epsilon, 1])\). Finally, in the case that the support is neither contained in a set of the form \((b, +\infty)\) nor \((-\infty, c)\), we can consider upper and lower tails in \(1\) by setting \(\{a_1,...,a_m\} =  F^{-1}((\epsilon, 1-\epsilon))\)

\section{Experiments}
We used Pytorch \cite{NEURIPS2019_9015} as our machine learning library for both the VAE and NAS experiments. For the VAE and NAS experiments, each individual run for a given seed used one GPU. We used Nvidia GeForce GTX 1080, and Nvidia Titan X GPUs on an internal cluster.

\subsection{Variational Auto-Encoder}
\label{appendix:vae}
The anonymous code for the VAE experiment, as well as all the plots in the paper, can be found at: \url{https://anonymous.4open.science/r/dbsurf_vae-3D9D/readme.md}

The encoder and decoder of the VAE both have 2 hidden layers of 200 units interspersed with Leakly ReLU \cite{leakyrelu} with a coefficient of 0.3, similar to \cite{disarm}. We used the Adam \cite{Kingma2013AutoEncodingVB} optimizer for both the encoder and decoder with a learning rate of $3\times10^{-4}$. The prior is drawn from a factorized Bernoulli distribution $p(z) \sim \text{Bernoulli}(0.5)$. The optimization is run for $1\times10^6$ steps with a batch size of 100. We pre-processed the data using dynamic binarization and by subtracting the mean of the dataset before passing it to the network. We ran the experiments for 5 seeds (0-4) across the three datasets MNIST, Omniglot, and FashiomMNIST.

\subsection{Neural Architecture Search}
\label{appendix:nas}
The anonymous code for the NAS experiment can be found at: \url{https://anonymous.4open.science/r/few_shot_nas-2657/README.md}

NasBench201 \cite{Dong2020NAS-Bench-201:} is a common benchmark for NAS algorithms. 
They create architectures by stacking search cells together to create a supernetwork. 
Each cell is a DAG with the edges representing operations and the nodes the intermediate representations.
On any given edge 5 operations can be selected $\{zero, skip-connection, 1\times1 conv, 3\times3 conv, 3\times3 avg pool\}$ and there are 4 nodes per cell, meaning a total of 6 edges.
The size of set $\mathcal{A}$ is $5^6 = 15,625$ architectures trained on CIFAR10, CIFAR100, and ImageNet16-120. We refer to their paper \cite{Dong2020NAS-Bench-201:} for more details about the search space. 

NasBench is a useful benchmark for NAS since the ground truth accuracy of the models is known. This means the process of searching for an architecture boils down to training an algorithm and then querying NasBench to see how good the architecture you found was. 

There are two phases to our algorithm, a warm-up phase and a training phase. In the warm-up phase, the architecture parameters are set as uniform and are not being updated. In this phase, we solely train the weights so that we are not updating the architecture parameters immediately from random initialization. Note that we are still using our procedure so we train the weights by sampling paths.

The second phase is the training phase and here we interleave one step of updating the weights with one step of updating the architecture parameters. We do not have separate train and validation sets and instead use the same samples to update both the weights and the architecture parameters as proposed in \cite{2020dsnas}. This means our technique, moves in the direction of not needing to be retrained after the architecture is found.

In the training phase, like \cite{2020dsnas} we perform pruning where we fix an operation on an edge if the difference between the weight of the best operation and the second best operation becomes larger than a threshold. For our experiments, this threshold is set to 0.8. 

For both phases, we use the entire training set with a batch size of 128 for CIFAR10 and CIFAR100 and a batch size of 512 for ImageNet16-120. For each batch of data, we sample $n = 5$ paths.

The \textbf{warm-up} stage lasts for 5 epochs and for the model weights we use SGD with momentum 0.9, weight decay of $4.0\times10^{-5}$ and a linear learning rate scheduler that starts at $2.0\times10^{-4}$ and ends at $1.0$. The loss for the model weights is the cross entropy loss with label smoothing of 0.1.

The \textbf{training} phase lasts 45 epochs and for the model weights, we use SGD with momentum 0.9, weight decay of $4.0\times10^{-5}$, and a Cosine Annealing learning rate scheduler starting from 0.5 and annealing to 0.0 over 235 epochs. The architecture parameters are updated using Adam with a learning rate of 0.001 and betas of [0.5, 0.999]. The loss for the architecture params is cross entropy loss with label smoothing of 0.1.

We ran the experiments for DBsurf on 4 seeds (0, 5, 10, 15) across the three datasets, CIFAR10, CIFAR100, and ImageNet16-120.
For the other techniques in the table we use the results as reported in \cite{hu2022generalizing}.

\section{Code Snippet}
\begin{lstlisting}
import numpy as np

def DBsample(probs, num_samples, alpha=1):
    """Discrepancy Based Sampling - Implemented for bernoulli distributions
    Args:
        probs (np.ndarray): Probability vector representing parameter for a 
        bernoulli distribution
        num_samples (int): Number of samples to generate
        alpha (float): bias / negative correlation parameter

    Returns:
        np.ndarray: Samples according to DBsample (num samples, dimension)
    """
    samples = []
    q = probs
    p_hat = probs
    for i in range(num_samples):
        u = np.random.uniform(size=[probs.shape[-1]])
        z = 1 * (u < q)
        samples.append(z)
        p_hat = ((i)*p_hat + z)/(i+1)
        q = np.clip(probs *  (1 + alpha) - alpha * p_hat,a_min=0, a_max=1)
    return np.array(samples)
\end{lstlisting}

\end{document}